\documentclass{article}
\usepackage{spconf,amsmath,graphicx,hyperref}
\usepackage{multirow}
\usepackage{subfigure}
\usepackage{graphicx}
\usepackage{subcaption}

\title{
Where, Not What: Compelling Video LLMs to Learn Geometric Causality for 3D-Grounding
}
\name{Yutong Zhong
\thanks{}
}
\address{New York University\\
         yz9760@nyu.edu}
\begin{document}
%
\maketitle
\begin{abstract}
Multimodal 3D grounding has garnered considerable interest in Vision-Language Models (VLMs) \cite{yin2025spatial} for advancing spatial reasoning in complex environments. However, these models suffer from a severe "2D semantic bias" that arises from over-reliance on 2D image features for coarse localization, largely disregarding 3D geometric inputs and resulting in suboptimal fusion performance. In this paper, we propose a novel training framework called What-Where Representation Re-Forming (W2R2) to tackle this issue via disentangled representation learning and targeted shortcut suppression. Our approach fundamentally reshapes the model's internal space by designating 2D features as semantic beacons for "What" identification and 3D features as spatial anchors for "Where" localization, enabling precise 3D grounding without modifying inference architecture. Key components include a dual-objective loss function with an Alignment Loss that supervises fused predictions using adapted cross-entropy for multimodal synergy, and a Pseudo-Label Loss that penalizes overly effective 2D-dominant pseudo-outputs via a margin-based mechanism. Experiments conducted on ScanRefer and ScanQA demonstrate the effectiveness of W2R2, with significant gains in localization accuracy and robustness, particularly in cluttered outdoor scenes.
\end{abstract}
\section{Introduction}
\label{sec:intro}
3D grounding matters, humans live in a 3D world and use natural language to interact with a 3D scene.\cite{qin2023langsplat} However, progress is constrained by data scarcity: state-of-the-art 3D models are “limited by datasets with a small number of annotated data.”\cite{xue2023ulip} To mitigate this, VG-LLM, recent systems pair a 2D encoder with a 3D geometry encoder, explicitly integrate[s] 3D visual geometry priors into MLLMs, processing images by “both a conventional visual encoder and the newly integrated 3D visual geometry encoder. \cite{zheng2025learning} A strong geometry backbone here is VGGT, which directly infers all key 3D attributes of scenes of its view.\cite{wang2025vggt} Another line aligns modalities via a shared embedding space: ULIP unifies image–text–point-cloud representations, ULIP-2 scales training by automatically generating holistic 3D-shape descriptions, Point-Bind aligns points with image/language/audio/video for broad transfer, OpenScene co-embeds 3D points with text and pixels in CLIP space, and LERF/LangSplat build 3D language fields for open-ended queries.\cite{xue2023ulip, xue2024ulip,  guo2023point, qin2023langsplat,kerr2023lerf, peng2023openscene}

Many claimed 3D benchmarks can be easily solved by applying a generic 2D VLM to rendered views indicating a very strong 2D bias.\cite{jin2025revisiting} Overreliance on 2D appearance / semantics when fusing with 3D cues has been documented as “2D-cheating”; object-centric, chain-of-analysis evaluations have been proposed, yet they remain evaluation level and do not disentangle 2D and 3D representations during training\cite{huang2025unveiling}

To solve this issue, we introduce W2R2, a multimodal 3D grounding training framework that changes internal representations without changing the inference architecture.
First, we explicitly mine a 2D-shortcut answer through the use of attention anchors that yield a coarse 2D-only localization.
Next, we introduce a Shortcut Suppression Loss that penalizes confident agreement with this shortcut and implicitly pushes the model away from the 2D route.
Finally, a Representation Re-Forming Loss enforces a what–where split, where 2D features carry semantics but 3D features define grounding. This keeps the shortcut semantically similar but spatially removed from the final answer and fosters reliance on 3D cues.

\begin{figure}[ht] 
    \centering
    \includegraphics[width=0.45\textwidth]{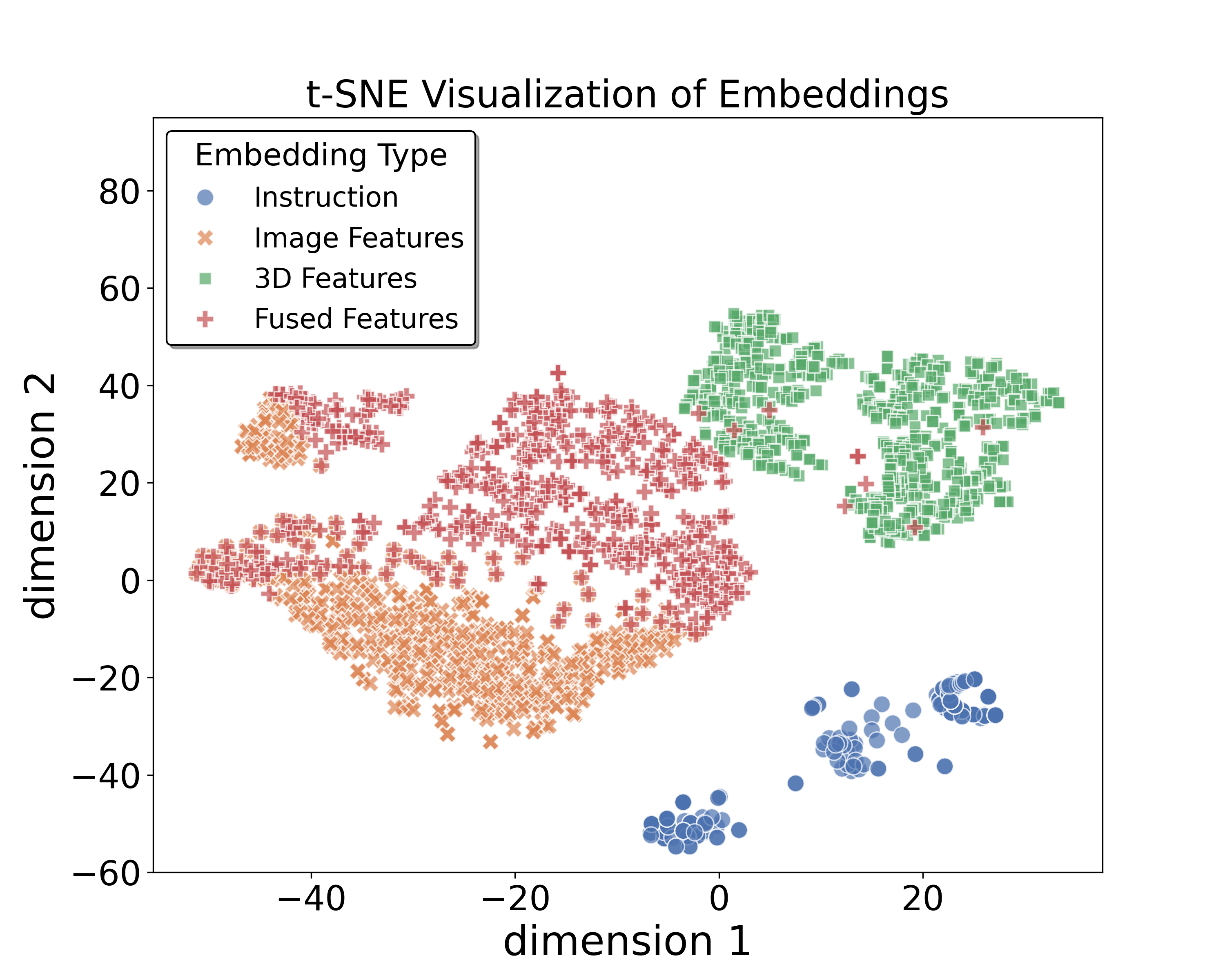}
    \caption{t-SNE visualization of feature distributions, highlighting a 2D semantic bias. The Fused Features show a much closer proximity to the 2D Image Features than to the 3D Features.}
    \label{fig:t-SNE}
\end{figure}
\section{Preliminary}

In contrast to 3D-LLMs that work directly with point clouds, our framework reconstructs 3D features from multi-view 2D images and combined features of the original 2D image with the reconstructed 3D geometric features is provided in the form of 3D grounding. Our method takes careful advantage of the strong pretrained semantic capacity of vision–language models (VLMs) on 2D image, but this leads to an inherent challenge that has yet to be fully resolved. The question is whether the model is actually learning to leverage the reconstructedz 3D geometric information or is acting on the "more familiar" 2D semantics as a quick escape.\cite{li2025does, huang2025surprise3d, deng2024can, qi2025beyond}We designed the following two diagnostic experiments to test this question.

First, we carried out a behavioral diagnostic to operationalize the theoretical reliance of the model on 3D features. We ablated the 3D input branch and examined localization on the ScanRefer dataset under the condition of using 2D image features-in-other-words-and-no-3D data. To our surprise, as shown in Table~\ref{table:table2}, modern VLMs have localization accuracy that is still clearly above a random-guess baseline despite having neither 2D nor 3D data. This indicated the presence of a ubiquitous 2D semantic shortcut in which models are relying on 2D semantics and are not actually doing 3D geometry awareness.

The model therefore learns to take advantage of strong 2D semantic information without developing a deep understanding of accurate 3D geometric relations. To explore the internal reason for this shortcut behavior, we examined the model’s latent representation space via t-SNE. Results from Figure~\ref{fig:t-SNE} establish a clear representation misalignment between modalities: 2D image features and corresponding reconstructed 3D features inhabit very different regions in latent space. Significantly, the fused representation collapses toward the 2D feature cluster, remaining far away from the clustered 3D features. This clustering results in the assumption that fusion has weight on 2D semantics over 3D geometry.
\begin{table}[t]
\centering
\caption{Diagnostic results on ScanRefer using only 2D inputs. All models achieve accuracy far exceeding random chance, indicating a heavy reliance on the 2D semantic features and a failure to effectively utilize 3D geometric features.}
\begin{tabular}{ccc}
\hline
\multirow{2}{*}{Method} & \multicolumn{2}{c}{ScanRefer} \\
\cline{2-3}
                        & Acc@0.25 & Acc@0.5 \\
\hline
\textit{Only 2D Input:}\\
SeqVLM                  & 47.8     & 39.3    \\
VG-LLM                  & 47.2     & 35.7    \\
VLM-Grounder            & 50.1     & 38.1    \\
\hline
\textit{2D + 3D Input:} \\
SeqVLM                  & 55.6     & 49.6   \\
VG-LLM                  & 53.2     & 49.7  \\
VLM-Grounder            & 62.4     & 53.2  \\
\hline
\end{tabular}
\label{table:table2}
\end{table}
Overall, these two experiments highlight the main issue of VLM-based 3D localization: there is an inherent 2D semantic bias. The evident “shortcut behavior” (Table~\ref{table:table2}) is a symptom of the misalignment of the representations (Figure~\ref{fig:t-SNE}). Since fusion occurs based on strong 2D features, the model does not have to really learn or be dependent on the more complicated 3D geometric evidence. This comprimised use of 3D evidence is an important barrier to higher accuracy 3D grounding. Therefore, the main goal of this work is to cut this bias and transform the representation space to allow for actual 3D scene understanding.
\section{PROPOSED METHOD}
\label{sec:method}
Instead of trying to eliminate 2D semantics, we tackle 2D shortcut learning in 3D LLMs by applying What–Where Representation Re-Forming (W2R2), a training scheme that reformulates internal representations. W2R2 uses two forward passes with shared parameters: full fused and 2D-only, and applies an objective to pull-push the model towards 3D-informed predictions, while at the same time pushing against overly strong 2D-only source predictions. The push-pull structure also decouples the representation space: the 2D features have semantics/coarse cues (what), and the 3D features already were dominating precise spatial grounding (where) for encouraging the model to use 3D geometry for grounding.

\begin{figure*}[ht]
    \centering
    \includegraphics[width=\textwidth]{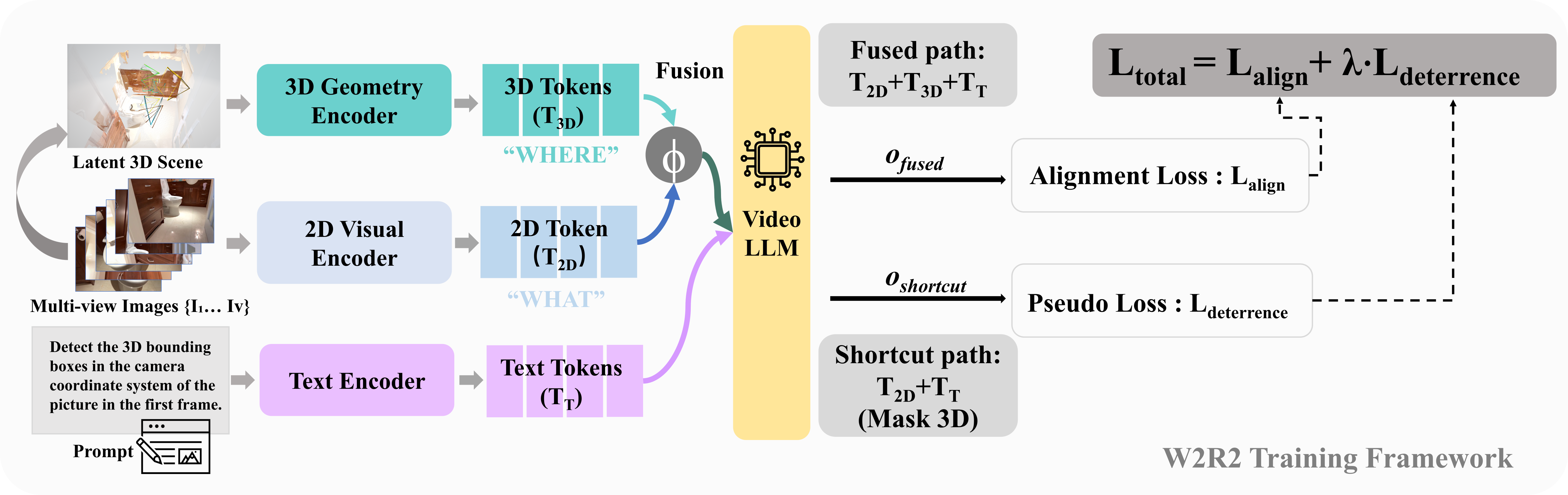} 
    \caption{Overview of our proposed W2R2 training framework.}
    \label{fig:framework}
\end{figure*}

\subsection{Baseline \& Notation}
Given multi-view RGB images $I=\{I_i\}_{i=1}^{V}$ with camera intrinsics/extrinsics $\{K_i,R_i,T_i\}_{i=1}^{V}$, a 3D encoder produces geometry-aware features $F_{\mathrm{3D}}=\mathcal{E}_{\mathrm{3D}}(I)$ \cite{rukhovich2022imvoxelnet}, while a 2D encoder provides semantic features $F_{\mathrm{2D}}=\mathcal{E}_{\mathrm{2D}}(I)$. We fuse the two streams via a generic operator $\Phi(\cdot,\cdot)$ (e.g., concatenation / cross-attention) and feed them with language query $q$ to a decoder/LLM:
\begin{equation}
o_{\text{fused}}=\mathcal{D}\big(\Phi(F_{\mathrm{2D}},F_{\mathrm{3D}}),\,q\big).
\end{equation}
We supervise the fused prediction with a standard alignment loss, while CE stands for cross-entropy
\begin{equation}
\mathcal{L}_{\text{align}}=\mathrm{CE}\big(o_{\text{fused}},\,y\big),
\end{equation}
\subsection{Formalizing the 2D Shortcut}
To expose the semantic shortcut, we ablate the 3D branch and define a 2D-only prediction:
\begin{equation}
o_{\text{short}}=\mathcal{D}\big(\Phi(F_{\mathrm{2D}},\mathbf{0}),\,q\big).
\end{equation}
This path captures coarse localization driven by strong 2D semantics and will be used for targeted regularization; gradients through this branch will be blocked in the push term.
\subsection{W2R2: Pull--Push Training}
\label{sec:w2r2}
At each iteration we run two forward passes with shared parameters $\theta$:
\begin{align}
o_{\text{fused}} &= f_{\theta}\big(\Phi(F_{\mathrm{2D}},F_{\mathrm{3D}}),\,q\big), \\
o_{\text{short}} &= f_{\theta}\big(\Phi(F_{\mathrm{2D}},\mathbf{0}),\,q\big).
\end{align}
W2R2 reshapes the representation space via a pull--push objective:
(i) \emph{Pull} aligns the fused output with the ground truth using $\mathcal{L}_{\text{align}}$.
(ii) \emph{Push} discourages over-reliance on the 2D path by penalizing a too-good 2D-only solution:
\begin{equation}
\mathcal{L}_{\text{deterrence}}
= \max\!\Big(0,\, s\big(\operatorname{loU3D}(o_{\text{short}}),\,y\big)-\mu\Big)
\end{equation}
where $s(\cdot,\cdot)$ is a task similarity (e.g., $\mathrm{IoU}_{3\mathrm{D}}$ for grounding), $\mu\in(0,1)$ is a tolerance margin, and $\operatorname{stopgrad}(\cdot)$ blocks gradients through the shortcut branch to prevent degenerate updates that only worsen the 2D path.
\subsection{Total Objective \& Effect}
The overall objective is
\begin{equation}
\mathcal{L}_{\text{total}}=\mathcal{L}_{\text{align}}+\lambda\,\mathcal{L}_{\text{deterrence}}, \qquad \lambda>0.
\end{equation}
The pull term rewards 3D-informed fused solutions, while the push term activates only when the 2D-only output is too close to the ground truth, thereby deterring shortcut reliance. This push--pull dynamics reshapes the representation space: 2D features retain semantics/coarse cues (\emph{what}), whereas 3D features dominate fine-grained spatial grounding (\emph{where}).
\begin{table*}[h]
\centering
\caption{Overall comparison with state-of-the-art methods on three benchmark datasets.}
\label{tab:main_results}
\begin{tabular}{lcccccc}
\hline
\multirow{2}{*}{Model} & \multicolumn{2}{c}{ScanRefer} & \multicolumn{2}{c}{Scan2Cap} & \multicolumn{1}{c}{ScanQA} & \multirow{2}{*}{Overall} \\
\cline{2-6}
\multicolumn{1}{c}{} & Acc@0.25 & Acc@0.5 & B-4@0.25 & C@0.25 & EM & \\
\hline
SPAR & 48.8 & 43.1 & - & - & - & - \\
SeqVLM & 55.6 & 49.6 & 41.1 & 82.3 & 30.2 & 51.8 \\
VLM-Grounder & 62.4 & 53.2 & 45.1 & 70.1 & 29.1 & 52.0 \\
SeeGround & 44.1 & 39.4 & 47.1 & 80.6 & 26.9 & 47.2 \\
Video-3D LLM & 58.1 & 51.7 & 41.3 & 83.8 & 30.1 & 53.0 \\
\hline
\textbf{W2R2 LLM} & \textbf{63.6} & \textbf{54.2} & 44.8 & 82.1 & \textbf{30.8} & \textbf{55.1} \\
\hline
\end{tabular}
\end{table*}
\section{Experiement}
\label{sec:experiement}
\subsection{Data and evaluation metrics}
\textbf{Dataset.}
We conduct experiments on three popular benchmarks for 3D grounding and captioning: ScanRefer\cite{10.1007/978-3-030-58565-5_13}, Scan2Cap\cite{chen2021scan2cap}, and ScanQA\cite{azuma2022scanqa}. For 3D visual grounding, we test our model on ScanRefer, which requires localizing unique objects in single-target scenarios. For dense captioning, we use the Scan2Cap benchmark, which involves generating descriptive captions for all objects in 3D scenes. For question answering, we employ ScanQA for spatial reasoning tasks. To perform evaluations, we follow prior works \cite{chen2020scanrefer} and use the validation of ScanRefer, Scan2Cap, and ScanQA.

\textbf{Metrics.} We adopt widely used evaluation metrics for each benchmark. For ScanRefer, we report threshold-based accuracy metrics, specifically Acc@0.25 and Acc@0.5, where a prediction is considered correct if its Intersection over Union (IoU) with the ground truth exceeds 0.25 and 0.5, respectively. For Scan2Cap, we apply CIDEr@0.5IoU and BLEU4@0.5IoU (denoted as C@0.5 and B-4@0.5), combining traditional image captioning metrics with the IoU between predicted and reference bounding boxes. For ScanQA , we use Exact Match accuracy (denoted as EM) to measure precise alignment with ground-truth answers.

\subsection{Main Results}
As shown in Table~\ref{tab:main_results}, our W2R2 framework achieves the best overall performance (\textbf{55.1}) across three benchmark tasks, demonstrating its superiority in complex 3D scene understanding. In the core 3D visual grounding task (ScanRefer), W2R2 sets a new state-of-the-art with \textbf{63.6\%} Acc@0.25 and \textbf{54.2\%} Acc@0.5, validating the effectiveness of suppressing 2D shortcuts to enhance 3D spatial awareness. Similarly, in 3D question answering (ScanQA), W2R2 leads with a \textbf{30.8\%} EM score, outperforming all competing methods. For 3D dense captioning (Scan2Cap), W2R2 delivers highly competitive results, comparable to leading approaches. These findings highlight that W2R2 not only significantly boosts localization accuracy but also maintains robust language generation, confirming the effectiveness and robustness of our Representation Re-Forming approach.

\subsection{Ablation Studies}
\label{sec:ablation_hyperparams}

\begin{figure}[] 
    \centering
    \includegraphics[width=0.23\textwidth]{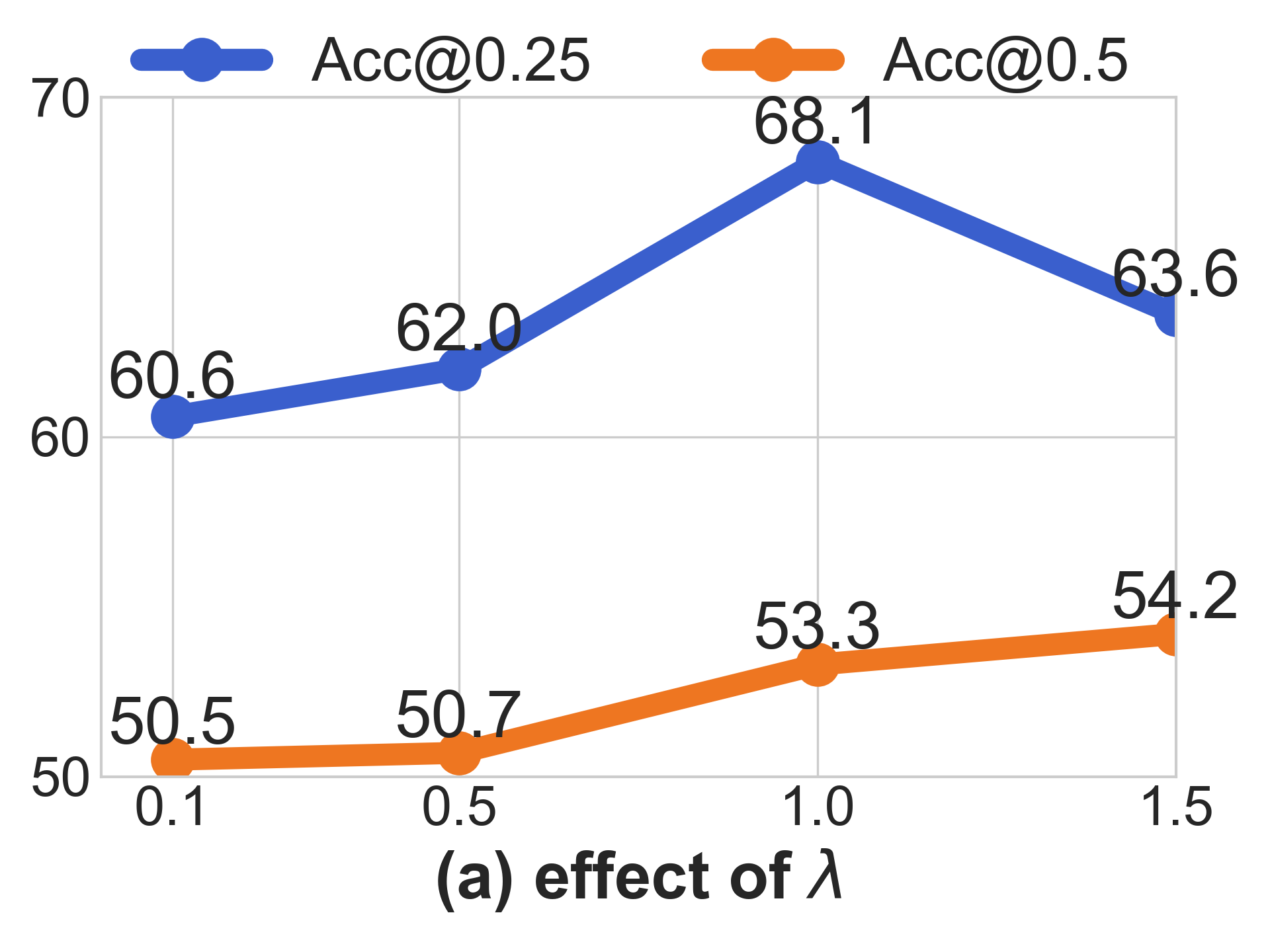}
    \includegraphics[width=0.23\textwidth]{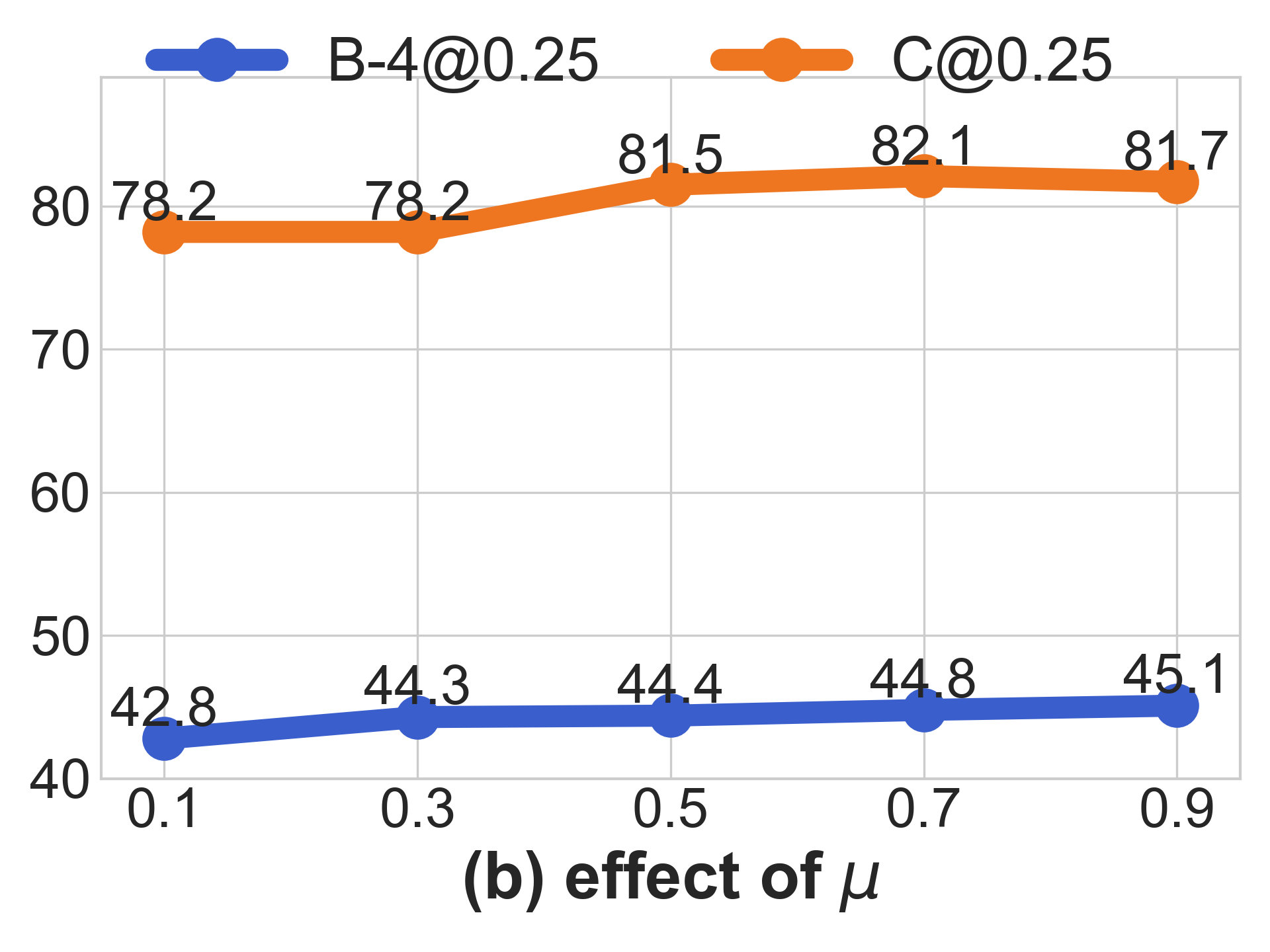}
    \caption{Ablation studies for hyper-parameters $\lambda$ and $\mu$.} 
    \label{fig:figure2}
\end{figure}

In this section, we analyze the impact of two key hyper-parameters in our framework: the suppression strength, $\lambda$, of the pseudo-path loss, and its activation threshold, $\mu$.
First, we investigate the effect of $\lambda$ on 3D grounding performance using the ScanRefer dataset. As shown in Figure~\ref{fig:figure2}(a), increasing $\lambda$ from 0.1 to 1.0 leads to a substantial improvement in the Acc@0.25 metric, which peaks at 68.1\%. Interestingly, the more stringent Acc@0.5 metric continues to benefit from stronger suppression, reaching its maximum of 54.2\% at $\lambda=1.5$. This suggests that a sufficiently high suppression strength ($\lambda \ge 1.0$) is crucial for achieving high-precision localization.
Next, we evaluate the activation threshold $\mu$ on the Scan2Cap dataset. The results in Figure~\ref{fig:figure2}(b) show that captioning quality, measured by B-4@0.25 and CIDEr (C@0.25), significantly improves as $\mu$ is increased from a low value of 0.1. The C@0.25 metric peaks at 82.1\% with $\mu=0.7$, while the B-4 score is maximized at 45.1\% with $\mu=0.9$. This validates our design rationale: an overly aggressive threshold (low $\mu$) is detrimental, whereas $\mu=0.7$ provides an optimal balance for activating the shortcut suppression.
Consequently, for all our main experiments, we set $\lambda=1.5$ to prioritize performance on the more challenging high-IoU metric, and $\mu=0.7$ as our default configuration.

\section{Conclusion}
\label{sec:conclusion}
In this paper, we address the prevalent 2D semantic bias in multimodal 3D grounding tasks by proposing the What-Where Representation Re-Forming (W2R2) framework. Departing from prior methods that optimize feature fusion, W2R2 fundamentally restructures the model's representation space to enforce a clear division: 2D features are directed toward semantic identification ("what"), while 3D features are prioritized for spatial localization ("where"). This approach mitigates over-reliance on 2D cues and enhances geometric utilization.
Empirical results on multiple benchmarks show substantial gains in grounding accuracy without additional inference costs, all while maintaining the model's original captioning performance. 
Beyond simple suppression, future research will investigate the potential for a symbiotic relationship between the 2D and 3D pathways. We aim to explore whether the coarse localization from the 2D shortcut can actively guide the 3D geometric reasoning, transforming the shortcut from a liability into a complementary signal for mutual enhancement.

\bibliographystyle{IEEEbib}
\bibliography{strings,refs}

\end{document}